\pdfoutput=1

\documentclass[11pt]{article}

\usepackage{acl}

\usepackage{times}
\usepackage{latexsym}

\usepackage[T1]{fontenc}

\usepackage[utf8]{inputenc}

\usepackage{microtype}

\usepackage{graphicx}
\usepackage{tabularx}
\usepackage{amsmath}
\usepackage{amssymb}
\usepackage{multirow}
\usepackage{makecell}
\usepackage{spverbatim}
\usepackage{fancyvrb,xcolor}

\usepackage{inconsolata}

\title{Towards Understanding Counseling Conversations: \\Domain Knowledge and Large Language Models}

\author{Younghun Lee$^{\dagger}$, Dan Goldwasser$^{\dagger}$, Laura Schwab Reese$^{\ddagger}$\\
       $^{\dagger}$Department of Computer Science\\
       $^{\ddagger}$Department of Public Health\\
       Purdue University \\
        \texttt{\{younghun,dgoldwas,lschwabr\}@purdue.edu}}
        
\begin{document}
\maketitle
\begin{abstract}
Understanding the dynamics of counseling conversations is an important task, yet it is a challenging NLP problem regardless of the recent advance of Transformer-based pre-trained language models. This paper proposes a systematic approach to examine the efficacy of domain knowledge and large language models (LLMs) in better representing conversations between a crisis counselor and a help seeker. We empirically show that state-of-the-art language models such as Transformer-based models and GPT models fail to predict the conversation outcome. To provide richer context to conversations, we incorporate human-annotated domain knowledge and LLM-generated features; simple integration of domain knowledge and LLM features improves the model performance by approximately 15$\%$. We argue that both domain knowledge and LLM-generated features can be exploited to better characterize counseling conversations when they are used as an additional context to conversations.
\end{abstract}

\section{Introduction}

Online counseling has become a more significant part of mental health services over the last couple of decades as younger generations feel more emotionally safe with digital communication \citep{murphy1998writing, king2006online}. Although building therapeutic relationships and social presence through written communication may exhibit significant challenges compared to in-person services \citep{king2006online, norwood2018working}, text or chat based counseling services are irreplaceable; nearly $50\%$ of the United States population reside in a mental health shortage area where there are less than two psychiatrists per 100,000 residents \citep{morales2020call, cheng2021addressing}.

The conversation dynamics and therapeutic relationship between mental health providers and clients have been actively studied in the health science field, mainly analyzing mutual trust \citep{torous2018empowering}, empathy \citep{nienhuis2018therapeutic}, social presence \citep{gunawardena1995social}, and rapport-building \citep{bantjes2022digital}. Despite its importance, there's relatively little work done in analyzing linguistic components of counseling conversations and characterizing them to better understand the conversation dynamics.

Throughout this research, we aim to propose a systematic approach to better characterize counseling conversations. We hypothesize that the current state-of-the-art language models contain insufficient knowledge of the counseling domain in their parameters. Motivated by existing works using external knowledge for solving tasks such as question answering \citep{ma-etal-2022-open}, commonsense reasoning \citep{schick2023toolformer}, and language generation \citep{peng2023check}, this paper studies whether additional knowledge helps characterize counseling conversations. We suggest two different ways of obtaining this additional knowledge: human annotation and large language model (LLM) prompting.

In this paper, we measure the level of understanding counseling conversations by predicting conversation outcomes, i.e., whether the help seeker would feel more positive after the conversation or not. We empirically show that Transformer-based classifiers as well as state-of-the-art LLMs exhibit sub-optimal performances despite their strong ability on many downstream tasks. The paper then describes how domain knowledge is obtained in order to further emphasize the counselor's strategic utterances and the help seeker's perspectives. We show that the additional knowledge helps pre-trained language models better fit the dataset and perform well in predicting the conversation outcomes---simple integration of the knowledge and feature ensembling improves the model performance by approximately $15\%$. We further analyze the efficacy of different features and explain how these features help classifiers better predict the outcome.

\textbf{Key Contributions}: To the best of our knowledge, this is the first attempt to exploit LLMs as a knowledge extractor to better characterize counseling conversations. With better prompting, we expect LLMs to generate more meaningful knowledge and explanations to assess the help seeker's perspectives. These knowledge-infused language models can be further used to generate evidence of how the conversation is going and how the help seekers may feel in real-time during the conversation, and ultimately assist human counselors in providing better counseling. 

\section{Counseling Conversation Analysis}
In chat based services for crisis counseling, a help seeker starts a session seeking help and a counselor replies to it. There are two speakers in these chat sessions, a help seeker and a counselor. Following previous works in analyzing such conversations \citep{sharma2020computational, grespan2023logic}, we aim to analyze counseling conversations by observing two different levels of features---utterance level features and session level features. Utterance level features examine the characteristics of conversation turns (i.e. messages), whereas session level features consider different aspects that can be found throughout the whole conversation.

\subsection{Problem Formulation}
\label{sec:training-pipeline}
One of the main goals of this research is to train a model that understands the conversation text between a counselor and a help seeker. Existing works on counseling conversations measure the level of language understanding by evaluating the quality of language generation; the models are trained with language model objectives and they generate the most likely utterance given a snippet of a conversation history. However, widely-used metrics for language generation such as ROUGE \citep{lin-2004-rouge} and BLEU \citep{papineni-etal-2002-bleu} do not accurately assess the model's language understanding in this domain because defining the correct utterance given the conversation context is unclear; given the same conversation context, both an empathetic text and a solution-driven text can be considered as a good response at the same time. Alternatively, language models can be evaluated by asking humans to choose better generations from different models. However, this does not guarantee fair evaluations because humans who evaluate generated texts cannot fully understand the help seekers' perspectives.

Thus in this paper, we use a more easy-to-understand feature to define the level of understanding. We choose the help seeker's post-conversation survey answer to a question, \textbf{\textit{``Do you feel more positive after this conversation?''}}, as an output of each conversation instance. We train the model to solve a classification task to predict whether the help seeker has become more positive after having a conversation session. 

Regardless of a simple classification pipeline, this is a challenging NLP task as it requires models to understand the context of a conversation session and to read between the lines to assess the help seekers' feelings throughout the conversation. The help seeker's perspectives on the counseling session can be affected by many factors such as their situations, needs, the type of abuse, the counselor's tone, rapport-building strategies, the solutions suggested by the counselor, etc. Moreover, help seekers rarely express their negative emotions about how the counselor is doing during the conversation (e.g. \textit{``You are not helping.''}). In most cases, the help seekers rather show their gratitude to the counselor as a courtesy (e.g. \textit{``Thanks for the help.''}), yet respond to the post-conversation survey that they don't feel more positive after the conversation. Thus the models need to analyze not only the direct meanings of what help seekers say, but also identify different aspects such as whether the help seekers' needs are met, if the solutions are specific to the help seekers' situations, whether the counselors express their empathy, etc.


\subsection{Human-annotated Domain Knowledge}
\label{sec:utterance-level-feature}
To better characterize the conversation and predict whether the help seeker has become more positive, we first obtain domain knowledge from human annotation. One of the main research questions we aim to solve in this paper is whether domain-specific knowledge helps understand counseling conversations. 
We qualitatively analyze around 200 counseling conversation sessions from The Childhelp National Child Abuse Hotline\footnote{https://childhelphotline.org} and annotate utterance level features with pre-defined counseling strategies; we focus on annotating utterances from the counselors and investigate the effects of counseling strategies on the help seekers.

Both inductive and deductive processes are used to explore the counseling strategies; the first draft of the feature set was based on existing conversations related to child maltreatment \citep{cash2020m, schwab2019child, schwab2022they}, then it was revised based on the content of the conversations. The overall feature development process follows the adaptation of grounded theory described by \citet{schreier2012qualitative}. The annotators identify patterns that are not covered by the features used in the first draft, then they discuss differences, refine the annotation framework, and apply the new features to small batches of the data (30 instances). By iteratively following this process, the annotators have come to identify various emotional attending strategies such as active listening \citep{ivey1992basic}, validation \citep{linehan1997validation}, unconditional positive regard \citep{wilkins2000unconditional}, and evaluation-based language \citep{brummelman2016praise}. After the inter-annotator agreement score reaches 95$\%$ in assessing the small batches, the annotators identify utterance level features for the rest of the data. 


\subsection{LLM-generated Features}
Recent studies show that LLMs can solve many different NLP tasks including summarization, classification, generation, and question answering \citep{chintagunta-etal-2021-medically, chiu2021detecting, goyal2022news, lee2022does, liu-etal-2022-makes}, suggesting these models are capable of understanding natural language and reasoning with world knowledge. As our task not only requires language understanding but also applying real-world knowledge, we aim to explore whether LLMs can comprehend counseling conversations and provide meaningful features that can later be used to characterize them. As we focus on obtaining utterance level features from human annotation, we put more emphasis on retrieving session level features and the help seekers' perspectives using LLMs.

It is also beneficial to study the role of LLMs in representing conversation text regarding training efficiency. Analyzing multi-turn conversations using Transformer-based models often encounters trade-offs between maximum token limits and model complexity; smaller models could easily reach their maximum token limits to encode the whole conversation text and bigger models like LongFormer \citep{beltagy2020longformer} require a larger number of training instances to fine-tune their parameters. LLM-generated features have the potential to replace the lengthy conversation text and ultimately help reduce possible issues in training, especially when the number of training instances is not large enough to tune a complex model.

\subsection{Data}
The data for this study comes from the text and chat channel of The Childhelp National Child Abuse Hotline. The crisis counselors are professionals with specialized training in hotline services and child maltreatment, rather than volunteers or peers like 7cups\footnote{https://www.7cups.com}, TalkLife\footnote{https://www.talklife.com}, or other mental health related online communities\footnote{https://www.reddit.com/r/depression/}. We gained access to de-identified transcripts and metadata that anonymized and normalized all names and street addresses which relieves ethical concerns.

This research studies two streams of data. $\mathcal{D}_{small}$ refers to the dataset we purposely select for annotating utterance level features. We select 236 conversation instances out of 1,153 total conversations recorded during July 2020. The selection criteria were designed to have a more diverse demographic background of the help seekers and more number of conversation sessions with valid post-conversation survey answers. We have another stream of data, $\mathcal{D}_{large}$, which includes additional conversation sessions from August 2021 to December 2022 where the help seekers provided valid post-conversation survey answers. The major difference between $\mathcal{D}_{small}$ and $\mathcal{D}_{large}$ is that the former has annotated utterance level features and demographically diverse distributions among help seekers, while the latter has more number of conversation sessions.

All counseling conversations are recorded in English. For $\mathcal{D}_{small}$, around $70\%$ of the help seeker was female, and $55\%$ of the help seeker was the maltreated child. About $60\%$ of the help seekers are younger than 17 years old. 

The annotation team includes one of the authors, a graduate research assistant, and two collaborators at Childhelp. The author is a family violence prevention researcher with a Ph.D. in public health and a Master of Arts in counseling. The author also has experience conducting qualitative analyses of written hotline transcripts. The graduate research assistant was a Master of Public Health student and had worked on the author's research team for three years. The research assistant had experience with qualitative child maltreatment research. The Childhelp collaborators have substantial experience in hotline counseling and leadership. One has a Master of Science in Counseling Psychology. The second has a Master of Science in Family and Human Development and a Master of Education in Guidance Counseling.

\begin{table}\footnotesize
\centering
\begin{tabularx}{1.00\columnwidth}{l|r}
\hline
\multicolumn{2}{c}{\textbf{$\mathcal{D}_{small}$}} \\
\hline
Number of sessions & 236\\
Class distribution (neg/neu/pos) & 31 / 104 / 101\\
Date range & 30 \\
Avg/Max number of tokens per session & 1,075 / 4,773\\
Avg/Max number of turns per session & 27 / 143\\
\makecell[l]{Avg/Max number of annotated utteran-\\ce level features per session}& 11 / 45\\

\hline
\hline
\multicolumn{2}{c}{\textbf{$\mathcal{D}_{large}$}} \\
\hline
Number of sessions & 1,469\\
Class distribution (neg/neu/pos) & 238 / 627 / 604\\
Date range & 300 \\
Avg/Max number of tokens per session & 1,034 / 5,253\\
Avg/Max number of turns per session& 26 / 234\\
\hline
\end{tabularx}
\caption{\label{tab:data-statistics}
Statistics of the two datasets. Only $\mathcal{D}_{small}$ contains human annotated utterance level features.
}
\end{table}

As mentioned in \ref{sec:training-pipeline}, we consider the help seekers' post-conversation survey answers as a class. We take the answer to a question, \textbf{\textit{``Do you feel more positive after this conversation?''}}, as output and discard instances where the help seekers answered `Prefer not to answer'. The remaining classes are `A lot (positive)', `A little (neutral)', and `Not at all (negative)'. Detailed statics of the datasets and the class distributions are described in Table \ref{tab:data-statistics}.

\section{Models}

We implement baseline models with the conversation text and integrate varying features to evaluate their efficacy.

\subsection{Baseline}
Baseline models are implemented to measure the difficulty of predicting conversation outcomes. In this setting, we only provide the conversation text between the counselor and the help seeker, and the model is trained to infer a conversation outcome (i.e. whether the help seeker has become more positive). 
Baseline models are pre-trained BERT-based sequence classifiers that are fine-tuned on the dataset. We implement BERT \citep{devlin-etal-2019-bert}, DistilBERT \citep{sanh2019distilbert}, and RoBERTa \citep{liu2019roberta} sequence classifiers from the huggingface distributions\footnote{https://huggingface.co/docs/transformers}.

The average number of tokens in a conversation session is over a thousand (see Table \ref{tab:data-statistics}), whereas the aforementioned pre-trained classifiers can encode up to 512 tokens. Thus we truncate the conversation text; the model takes the first and the last $k$-turns of the conversation\footnote{We compare this method to other alternatives such as using LongFormer or LSTM-based models, yet truncation works the best.}. In general, the beginning of the conversation includes the reason why the help seeker reached out, and the conversation develops into solutions and suggestions towards the end of the conversation. From this observation, we hypothesize that the beginning and the end of the conversation can better characterize the content rather than letting the model encode the text from the beginning and truncate the rest of the text when it reaches the maximum token limits. We have experimented with different encoding approaches to test the hypothesis and found out that our encoding approach (i.e. using the first and the last $k$-turns) outperforms the plain encoding approach (i.e. encoding from the beginning until the token limit) by 4$\sim$9\% in macro F1 score.

Another baseline model we evaluate is the state-of-the-art LLMs. We prompt ChatGPT\footnote{We use version gpt-3.5-turbo-0613} in a zero-shot setting to predict the conversation outcome. Unlike BERT-based classifiers, ChatGPT can take up to 4,096 tokens and there are less than 10 instances that exceed this limit in the dataset. Thus in using ChatGPT, we only remove a couple of utterances for the conversation sessions exceeding the maximum token limit and use the whole conversation for the rest of the sessions. 

\subsection{Integrating Utterance-level Features}
\label{experiments-domain-knowledge}

Counseling strategies (i.e. utterance level features) are annotated for only a partial amount (i.e. $\mathcal{D}_{small}$) of the full dataset (i.e. $\mathcal{D}_{all} = \mathcal{D}_{small} \cup \mathcal{D}_{large}$). To fully integrate utterance level features into conversation text, we implement simple classifiers that identify strategies in a counselor's utterance. Given a counselor's utterance and its previous $k$-turns of the conversation, classifiers assign correct utterance level features. Note that this is a multi-label classification as a counselor's utterance can exhibit multiple strategies at the same time.

\begin{table}\footnotesize
\centering
\begin{tabularx}{0.96\columnwidth}{l|c}
\hline
\multicolumn{2}{c}{\textbf{Utterance-level Feature Prediction}} \\
\hline \hline
\textit{Fine-grained Feature Classification} &  \textit{F1}\\
\hline
BERT-based end-to-end classifier & 55.03\\
BERT-based 2-step hierarchical classifier & \textbf{56.49}\\
text-davinci-003, few-shot (2 samples) prompt & 48.87\\
text-davinci-003, few-shot (3 samples) prompt & 56.2\\
\hline
\hline
\textit{Grouped Feature Classification} &  \textit{F1}\\
\hline
BERT-based end-to-end classifier & \textbf{69.22}\\
text-davinci-003, few-shot (3 samples) prompt & 61.12\\
\hline
\end{tabularx}
\caption{\label{tab:strategy-clf-results}
Utterance level feature prediction results of BERT-based classifiers and LLM-based classifiers. Fine-grained feature classification models infer among 18 classes while grouped feature classification models assign classes from the grouped features (4 classes).
}
\end{table}
There are 18 distinct features identified from the annotation framework described in \ref{sec:utterance-level-feature}, yet we categorize them into 4 groups, \textit{`Emotional Attending'}, \textit{`Fact Related'}, \textit{`Problem Solving'}, and \textit{`Resources'}. The performance of different classifiers in predicting utterance level features in Table \ref{tab:strategy-clf-results} shows trade-offs between the feature's expressibility and the model's faithfulness; when a more fine-grained set of features is used, more diverse utterance level information is added but the accuracy of the inferred features from the classifier is likely to be lower. Given the classification results, we choose to use groups of features for weak supervision. More details of the features and how they are grouped are described in Appendix \ref{sec:appendix-utterance-coding}.

{\renewcommand{\arraystretch}{1.1}
\begin{table*}\scriptsize
\centering
\begin{tabularx}{0.95\textwidth}{l|X}
\hline
\textbf{Prompt Type} & \textbf{Feature Examples} \\
\hline
Help seeker's identity & $\{$Maltreated child, Family member, Peer/Friend, Other known adult, Unknown person, Other$\}$\\
Perpetrator's identity & $\{$Parents, Siblings, Step-parents, Ex-partners, Other family member, Peer/Friend, Other$\}$\\
Type of abuse & $\{$Physical, Verbal/Emotional, Neglect/Careless, Stress from family/friends/school$\}$\\
Severity of abuse & $\{$Imminent danger, Persistent abuse, Poor care, Casual behavior$\}$\\
\hline
Help seeker's needs & $\{$Seeking resources, Getting emotional support, Reporting the situation, Practical advice, Not clear$\}$\\
Counselor's response & $\{$Providing resources, Reflection of feelings, Affirmation or reassurance, Providing advice, Not clear$\}$\\
Counselor's strategies & $\{$Interpreting, Reflecting feelings, Asking questions, Validating, Providing information$\}$\\
\hline
What's been tried & $\{$Contacting authorities, Talking to professionals, Talking to others, Self care methods, Others, None$\}$\\
Counselor's advice & $\{$Contacting authorities, Talking to professionals, Talking to others, Self care methods, Others$\}$\\
Help seeker's reaction& $\{$Accepting, Accepting with concern, Doubting, Has already been tried, Denying$\}$\\
\hline
Counselor's negative attitudes & $\{$Trivializing issues, Lacking validation, Pushy tone, Lacking exploration, Lacking solutions$\}$\\
Help seeker's negative attitudes & $\{$Yes, No$\}$\\
\hline
\end{tabularx}
\caption{\label{tab:llm-features-design}
Main features we aim to retrieve from LLMs. Detailed design of each prompt is described in Appendix \ref{sec:llm-prompts-details}}
\end{table*}
}

Using the BERT-based classifier for grouped utterance features, we automatically annotate the counselors' utterances that are not annotated by humans (i.e. $\mathcal{D}_{large}$). In order to better represent the conversation text, we integrate utterance level features into the existing text data. Specifically, we add this additional knowledge as special tokens that further explain the message that follows. Refer to a short snippet of a conversation and the same conversation with utterance level features added, for instance.
\begin{Verbatim}[fontsize=\small,commandchars=\\\(\)]
[Original Conversation]
Help seeker: I am abused by my parents.
Counselor: I am sorry that happened.

[Conversation with Utterance Features]
Help seeker: I am abused by my parents.
Counselor: (\color(red)<Emotional Attending>) I am sorry that
happened.
\end{Verbatim}

Using the inputs with utterance feature addition, we train BERT-based classifiers to predict conversation outcomes and compare their performance with the baseline models.

\subsection{Extracting Session-level Features using LLMs}
\label{sec:llm-feature-extraction}
The main advantages of using LLMs to extract relevant features from conversation text are two-fold: compressing lengthy conversation text, and cost efficiency. When LLM-generated features exhibit representation power comparable to the original conversation text, we can compress the lengthy conversation input by replacing it with LLM-generated features. Also, annotating domain knowledge following the process we perform in \ref{sec:utterance-level-feature} is costly and time-consuming, thus it would be cost efficient if LLMs are able to provide useful knowledge to characterize conversation text without having human annotators trained to analyze the data.

We first evaluate an LLM's ability to predict utterance level features. Table \ref{tab:strategy-clf-results} illustrates that the performance of prompting the text-davinci-003\footnote{We use the largest model at the time of running experiments. Note that the results might change with the most recent models.} model in both zero-shot and few-shot settings is worse than BERT-based classifiers. From the observation, we hypothesize that identifying utterance level features from the conversation is highly contextual and it requires fine-tuning rather than prompting LLMs. Thus we focus on retrieving session level features that are less contextual but meaningful in order to better understand the help seekers' perspectives.

We design 12 questions that cover a sufficient range of understanding how the conversation went and what the help seeker would have thought, and prompt ChatGPT in a zero-shot setting to get the answers to the questions. The questions focus on analyzing the help seekers' needs, the corresponding solutions suggested by the counselors, and also observe both of their attitudes. We consider the answers generated from these questions as session level features as they need to be answered by reading the whole conversation text. To alleviate the issues of providing generic answers or being hallucinated, we force ChatGPT to answer the questions by selecting from pre-defined choices. We have 60 choices (i.e. features) in total and Table \ref{tab:llm-features-design} shows examples of the questions and their corresponding features.

Having features selected by ChatGPT, we first process them as one-hot vectors and train machine learning models to predict conversation outcomes. Various models including Logistic Regression, Support Vector Classifier, Gaussian Naïve Bayes, and ensemble models such as Random Forest \citep{ho1995random} and AdaBoost \citep{freund1997decision} are implemented. 

Another way to utilize the session level features is to express them as a natural language explanation and encode them with BERT-based models. The following paragraph illustrates an example. 
\begin{Verbatim}[fontsize=\small,commandchars=\\\(\)]
[LLM-generated Features]
Help seeker's identity: Maltreated child
Perpetrator's identity: Parents
Type of abuse: Physical
...

[Natural Language Explanation of Features]
A (\color(red)maltreated child) has been experiencing (\color(red)physical)
abuse by their (\color(red)parents)...
\end{Verbatim}

One of the advantages of this approach is that these textualized features can be added to the conversation text and provide more parameterized information when BERT-based classifiers are trained. We concatenate the last hidden state representation of the two inputs (i.e. conversation text and session feature text) and train a classifier.

\subsection{Free-form LLM Generation}
In order to examine the efficacy of asking pre-defined questions in characterizing counseling conversations, we compare the features generated in \ref{sec:llm-feature-extraction} with free-form generation from LLMs. Instead of asking specific questions, we simply ask the ChatGPT model to summarize the conversation. We obtain two different summaries; one generates a plain summary, and the other is prompted to generate summaries, \textit{focusing on whether the help seeker would have felt more positive after the conversation}. The former contains information about the conversation only, while the latter includes ChatGPT's stance on whether the conversation affected the help seeker in a more positive way. When the summary is fed into the model with conversation text, the last hidden state of summary text from a BERT encoder is concatenated.

\section{Experimental Settings}
Very little difference exists between `positive' and `neutral' conversation outcomes. We combine these two classes and make the task as a binary classification task (i.e. `negative' v. `non-negative'). To evaluate and compare different models, we compute macro F1 scores and the recall values of the minority class (i.e. `negative' class). Models can achieve a satisfactory macro F1 score by minimally assigning minority class to test instances. In such cases, these models will score low recall on the minority class. However, models with higher recall on the `negative' class are more desirable in a real use case, as they identify more instances where the help seekers do not feel positive, and one can further assess what can be done alternatively.

The reported results are from DistilBERT-base-uncased classifier which works the best among all BERT based classifiers we implemented. Conversation text includes $k=4$ turns in the beginning and the end. We use the union of $\mathcal{D}_{small}$ and $\mathcal{D}_{large}$ as our main dataset, $\mathcal{D}_{all}$, with 60/20/20 splits of training, evaluation, and testing sets. All models are experimented with 10-fold cross validation. 

Table \ref{tab:experimental-results} illustrates the conversation outcome prediction results of various models and inputs. In the table, inputs are abbreviated as follows: \verb|Conv| is conversation text, \verb|Utter| means utterance level features are added to the conversation text, \verb|Session| is natural language explanation of ChatGPT generations about session level features, \verb|Summary| means plain summaries generated from ChatGPT, and \verb|Stance| is ChatGPT's summary with a stance on whether the help seeker feels positive or not.

\begin{table}\footnotesize
\centering
\begin{tabularx}{1.00\columnwidth}{l|c|c}
\hline
\multicolumn{3}{c}{\textbf{Conversation Outcome Prediction}} \\
\hline \hline
\textbf{Input $\Rightarrow$ Model} &  \textbf{F1} & \textbf{Recall}\\
\hline
\multicolumn{3}{c}{\textit{Baseline Models}}\\
\hline
\verb|Conv| $\Rightarrow$ DistilBERT & 61.91 & 31.39\\
\verb|Conv| $\Rightarrow$ ChatGPT & 63.23 & 25.28\\
\hline
\multicolumn{3}{c}{\textit{Utterance-level Features}}\\
\hline
$\pmb{\star}$ \verb|Utter| $\Rightarrow$ DistilBERT & 62.84 & 37.04\\
\verb|Utter| $\Rightarrow$ ChatGPT & 62.09 & 24.39\\
\hline
\multicolumn{3}{c}{\textit{Session-level Features}}\\
\hline
\verb|Session| one-hot vector $\Rightarrow$ AdaBoost & 63.84 & 24.82\\
\verb|Session| $\Rightarrow$ DistilBERT & 63.80 & 27.37\\
\verb|Conv|+\verb|Session| $\Rightarrow$ DistilBERT & 63.97 & 30.11\\
$\pmb{\star}$ \verb|Utter|+\verb|Session| $\Rightarrow$ DistilBERT & 64.60& 41.24\\
\hline
\multicolumn{3}{c}{\textit{Features from Summaries}}\\
\hline
\verb|Summary| $\Rightarrow$ DistilBERT & 62.36 & 29.56\\
\verb|Utter|+\verb|Summary| $\Rightarrow$ DistilBERT & 65.53 & 32.85\\
\fontsize{8}{10}{\verb|Utter|+\verb|Session|+\verb|Summary| $\Rightarrow$ DistilBERT} & 65.32 & 41.06\\
\verb|Stance| $\Rightarrow$ DistilBERT & 68.46 & 37.59\\
$\pmb{\star}$ \verb|Utter|+\verb|Stance| $\Rightarrow$ DistilBERT & 69.88 & 41.42\\
\fontsize{8}{10}{\verb|Utter|+\verb|Session|+\verb|Stance| $\Rightarrow$ DistilBERT} & 66.88 & 36.50\\
\hline
\multicolumn{3}{c}{\textit{Feature Ensembling}}\\
\hline
\makecell[l]{$\pmb{\star}$ \verb|Utter|+\verb|Session|+\verb|Summary|\\+\verb|Stance| $\Rightarrow$ Ensemble} & \textbf{71.29} & \textbf{49.27}\\
\hline
\end{tabularx}
\caption{\label{tab:experimental-results}
Macro F1 scores and recall values of the `negative' class. The input to the AdaBoost models are one-hot encoded vectors of session level features, and all other DistilBERT models get text inputs. Ensemble model stacks logits from different classifiers and learn a final Logistic Regression classifier. A leading star sign indicates the model with the best F1 and recall score within the same category.
}
\end{table}

\section{Discussion}
In this section, we further diagnose the model outputs and their relatedness to the features.

\subsection{Model Performance}
We empirically show that predicting the conversation outcome is not a trivial task regardless of its simple training pipelines. The first two rows in Table \ref{tab:experimental-results} show that the baseline models lack in performance. Although the ChatGPT model scores a higher macro F1 score, its low recall implies that the model predicts fewer conversation instances as `negative'. This validates our argument described in \ref{sec:training-pipeline}; predicting the conversation outcome is a challenging task and it requires more domain-specific knowledge rather than relying on the knowledge encoded in language model parameters.

Overall, the performance of language models incrementally improves by adding more features---utterance level features, session level features, and features from summaries---except for the case where \verb|Utter|+\verb|Stance| shows better performance than \verb|Utter|+\verb|Session|+\verb|Stance|. While the efficacy of session level features is not clear when it is used with summaries with stance, it helps the language model better perform when used with other features. Ensembling classifiers trained with different features not only mitigates the potential class imbalance issues but also produces the best F1 and recall scores.

\subsection{Effectiveness of Utterance-level Features}
Utterance level features can enhance the model's accuracy in general as well as its ability to identify `negative' class instances. Simple integration of utterance level features to the conversation (i.e. \verb|Utter|) improves the F1 score by $1.5\%$ and minority class recall by $18\%$ compared to the original conversation (i.e. \verb|Conv|). We observe that utterance level features also improve when both conversation text and session level features are used together; \verb|Utter|+\verb|Session| enhances the minority class recall by $37\%$ than \verb|Conv|+\verb|Session|, while maintaining F1 scores.

\begin{figure*}[t!]
    \centering
    \includegraphics[width=1.00\textwidth]{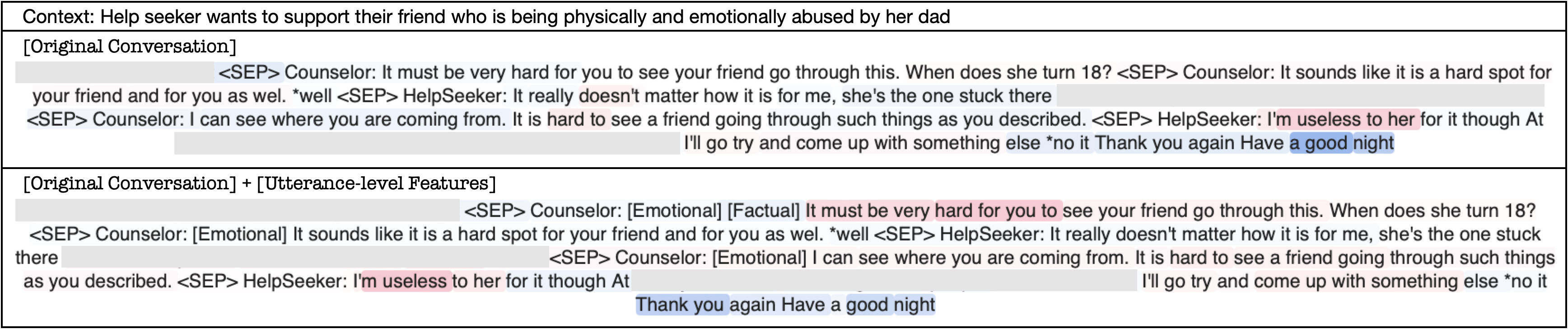}
    \caption{Shapley value of phrases in the counseling conversation (upper) and the conversation with utterance level features (lower). Highlighted area in red contributes the models to predict `negative' class, and area in blue contributes the opposite.}
    \label{fig:utterance-feature-explanation}
\end{figure*}

We compute the Shapley values and observe how utterance level features contribute differently to the classifier following the approaches proposed in SHAP \citep{lundberg2017unified}. Compared to the original conversation input, utterances that are integrated with features tend to contribute more to the inference, which potentially leads models to identify more `negative' instances. For instance, the counselor's utterance, \textit{``It must be very hard for you to ...''} in Figure \ref{fig:utterance-feature-explanation} contributes more to the final prediction when it appears with the utterance feature indicators, and it ultimately leads the model to infer a correct class, `negative'.


\begin{figure}[t!]
    \centering
    \includegraphics[width=1.00\columnwidth]{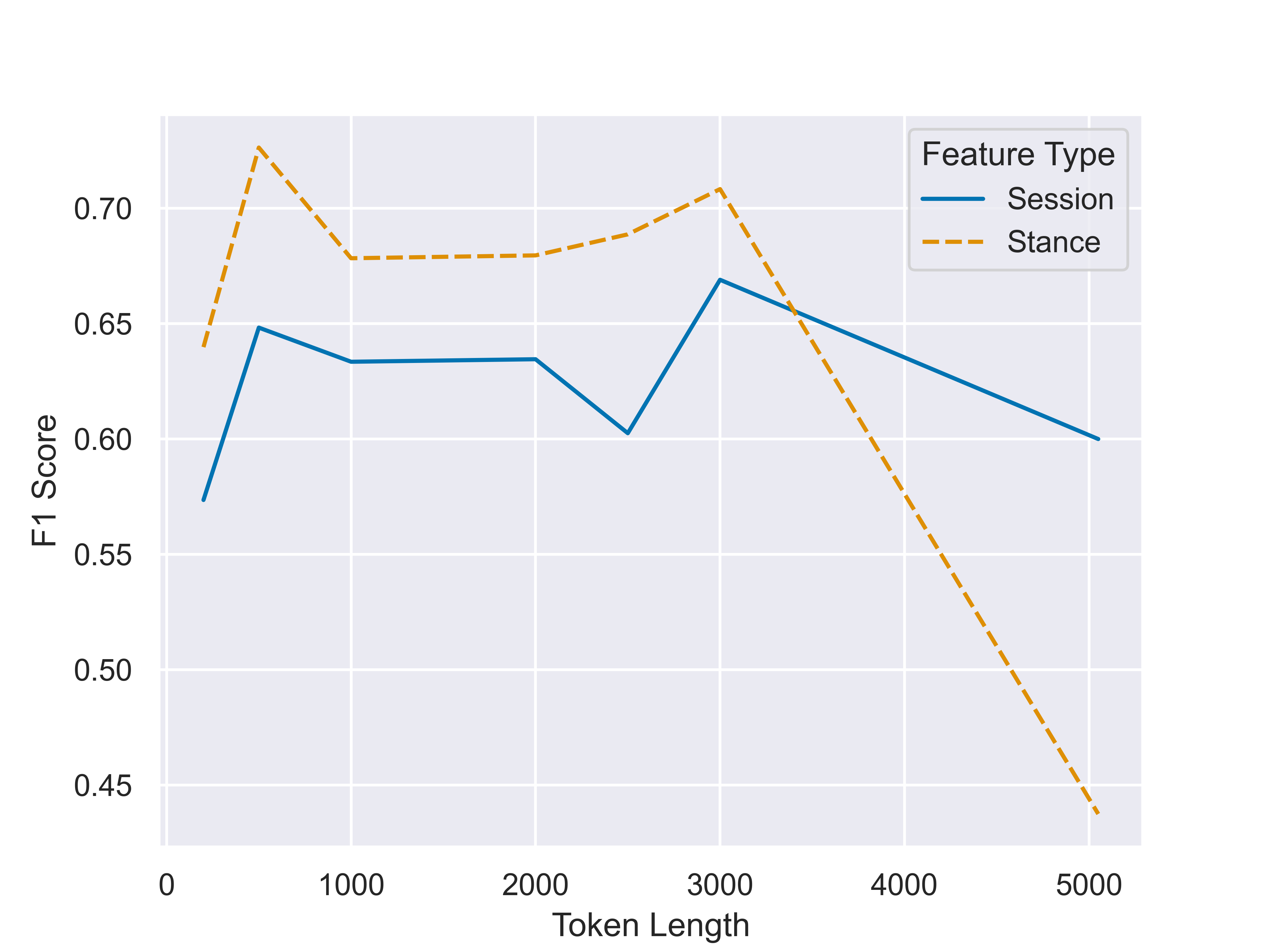}
    \caption{F1 score comparison between session level feature input and summaries with stance. Performance of summary with stance decreases when the length of the counseling conversation exceeds 3K tokens, while session level feature input shows more consistent performance.}
    \label{fig:downstream-task}
\end{figure}

\subsection{Effectiveness of Session-level Features}
Session level features show sufficient representation abilities compared to the original conversation text. Using session level features, either one-hot encoded or represented by BERT-based encoders, shows better performance in predicting the outcome even without considering the original conversation text. 

The effectiveness of session level features is arguable when it is used with features from summaries. While session level features improve the minority class recall for the plain summary features, summaries with stance can perform best without having session level features at all. This observation raises a question, \textit{``Are session level features essential when we have summaries with stance?''}. 

We further diagnose the performance of the two models, one using session level features and the other using features with stance with respect to the length of the conversation text. When the context is lengthy, we hypothesize that LLMs are susceptible to having more insufficient or incorrect generations in producing general summaries, compared to answering questions focusing on specific aspects. Figure \ref{fig:downstream-task} shows the F1 score of the two models with respect to the length of the conversation. As the conversation gets longer than 3K tokens, the performance of summaries with stance decreases while session level feature input shows consistency. This implies that obtaining summaries and using them as features becomes less consistent when the input conversation is lengthy, thus using session level features is more beneficial.

\subsection{Plain Summary v. Summary with Stance}
The difference between generating plain summary and summary with stance is very minimal in the prompts, yet their effectiveness varies significantly; using \verb|Stance| improves the macro F1 by $12\%$ and the minority class recall by $27\%$, compared to using \verb|Summary|. 
To further examine the commonalities and differences of the summaries generated by the two approaches, we identify distinct aspects that are captured in the summaries through clustering.

We split the summaries into sentences and run k-means clustering to group similar sentences together. Qualitative analysis shows that the plain summary generates more sentences mentioning the help seeker expressing gratitude at the end of the conversation, while the summary with stance generates whether the help seeker would feel more positive after the conversation. We argue that this difference leads the \verb|Summary| model to have a low recall on the `negative' class; having a summary sentence about the help seeker being thankful makes the classifier more likely to infer an instance as `positive', yet the expression of gratitude should not be considered as a significant feature as described in \ref{sec:training-pipeline}. Another difference is that the plain summary generates more details of the help seekers' situations, particularly about their parents being abusive, while the summary with stance focuses more on whether the counselor empathizes with the help seeker's situation. Figure \ref{fig:summary-clusters} in Appendix \ref{sec:appendix-clustering} illustrates the clustered sentences in the summaries and co-occurring themes in each cluster.

\section{Related Work} Several recent NLP works looked at analyzing counseling conversations and predicting their outcomes~\cite{althoff-etal-2016-large,perez-rosas-etal-2018-analyzing,perez-rosas-etal-2019-makes, grespan2023logic, li2023understanding}.
Similar to our approach, several work relied on domain knowledge to identify counseling strategies and conversational actions~\cite{lee-etal-2019-identifying,park-etal-2019-conversation,cao-etal-2019-observing}. For example, \citet{cao2019observing} employed behavioral codes of clients and therapists to provide real-time feedback to a therapist about the category of the current utterance and suggest the next category to apply.

Other works analyzed the conversational style of counselors, how it changes over time~\cite{zhang-etal-2019-finding,zhang-danescu-niculescu-mizil-2020-balancing} and the emotional support they provide~\cite{perez-rosas-etal-2017-understanding,sharma2020computational}. For example, \citet{sharma2020computational} proposed an empathy-based approach in understanding counseling conversations between a help seeker and peer supporters on TalkLife and \verb|r/depression| subreddits \citep{sharma2018mental}. 
\citet{liu2021towards} worked on guiding dialog models with emotional support strategy chains using 7cups dataset \citep{baumel2015online}. The authors evaluated the framework on BlenderBot \citep{roller2021recipes} and DialoGPT \citep{zhang2020dialogpt}. 

As counseling conversation analysis has been improving with the help of more representative language models over time, our research poses the initial attempt to utilize LLMs for reasoning about features relevant to conversational dynamics, and their relatedness to conversation outcomes.

\section{Conclusion}
We study the dynamics of conversations between crisis counselors and help seekers. 
Transformer-based models and the ChatGPT fail to predict whether the help seeker feels positive after the conversation. To better characterize counseling conversations, we integrate domain-specific knowledge, human-annotated utterance level features identifying counseling strategies, and LLM generated session level features portraying help seekers' perspectives. We show that ensembling additional features improves performance in predicting conversation outcomes. Analyses suggest that the features lead the model to focus more on the counselor's strategy-related utterances, and better represent lengthy conversations with session level features.

\section*{Limitations}
This paper shows the effectiveness of domain-specific knowledge and LLM generations in understanding counseling conversations. One of the major limitations of this work is the sub-optimal performance of LLM generated features. LLMs show great performances in many downstream tasks, especially when prompted with additional knowledge. Studying more approaches in prompt engineering to get more meaningful session level features with the help of human annotated features would be beneficial. Additionally, evaluating the quality of LLM generated features would improve the effectiveness of the features.

We did not fully explore the most efficient model structure to combine utterance level features and session level features. Multi-task learning objectives for utterance level features and session level features to be benefited from each other used in \citet{grespan2023logic} can be a future work we can consider. 

Another approach is to minimize the use of LLMs and train a model to generate features. One of the future approaches can be adopting the On Policy Learning framework and training a tunable language model, such as FLAN-T5 \citep{chung2022scaling}, to generate session level features given a conversation, that maximizes the rewards (i.e. the outcome prediction performance).

The effectiveness of the domain knowledge in understanding counseling conversations was shown in one data source. Due to their sensitivity, access to such conversation is often limited, and experimenting with additional datasets would help demonstrate the generalizability of our approach.

\section*{Ethics Statement}
To the best of our knowledge, this work has not violated any code of ethics. As the data of this research includes human subjects and their behaviors, this research has been approved by the Institutional Review Board. The annotators as well as the researchers signed data confidentiality agreements and received an online education regarding ethical guidelines. The personal information of help seekers, such as names and street addresses, is anonymized and normalized prior to the researchers obtaining the data. Sample conversations described in \ref{sec:utterance-level-feature} and \ref{sec:llm-feature-extraction} are synthetic examples. This paper illustrates a real example of a conversation snippet in Figure \ref{fig:utterance-feature-explanation}. We replace the details of the conversation with `Context', and erased some parts from the help seeker's utterances that are unnecessary in evaluating the models. We provide the code for future reproducibility of the work. The data will not be publicly shared or posted anywhere.

\section*{Acknowledgments}
This project is mainly supported by the Children's Bureau (CB), Administration for Children and Families (ACF) of the US Department of Health and Human Services (HHS) as part of a financial assistance award in the amount of \$6 million with 100\% percent funded by CB/ACF/HHS, and partially funded by NSF IIS-2048001 and DARPA CCU program. The contents are those of the author(s) and do not necessarily represent the official views of, nor an endorsement by, CB/ACF/HHS/DARPA, or the US Government. For more information, please visit Administrative and National Policy Requirements.


\bibliography{custom}

\appendix


{\renewcommand{\arraystretch}{1.1}
\begin{table*}[t!]\scriptsize
\centering
\begin{tabularx}{0.95\textwidth}{c|c|X}
\hline
\textbf{Abstract Category} &  \textbf{Feature} & \textbf{Description}\\
\hline
\multirow{13}*{Emotional Attending}  & Paraphrasing & Repeats what was said by the help seeker in a way that hones the focus of the conversation.\\
 & Interpreting & Offers a coherent overview of the situation and a supports the help seeker to see new patterns or ideas.\\
 & Reflecting feelings & Distills the help seeker's feelings to support in identifying what is most bothering them about the situation.\\
 & Validating & Affirms the help seeker, their feelings, and their thoughts to ensure that they are important.\\
 & Unconditional positive regard & Provides support of the help seeker, regardless of their behavior or things that have been done to them.\\
 & Open questions & Invites the help seeker to share about the experience that helps exploring the issues and eliciting details. \\
 & Praise & Approves the help seeker or their behavior.\\
 & Apology & Apologizes about technical difficulties or expresses their compassion for the help seeker and their situations.\\
\hline
\multirow{2}*{Fact Related} & Fact seeking & Asks questions about specific situations to get better understandings\\
 & Fact giving & Provides factual knowledge based on the help seeker's questions or their situations\\
\hline
\multirow{5}*{Problem Solving} & Asks what has been tried & Asks help seeker what they have tried to resolve the issue\\
 & Asks about supports/resources & Asks help seeker which  resources they tried or considered trying\\
 & Advice/idea giving & Suggests solutions to resolve the help seeker's issues\\
 & Pushes advice/resources & Continuously mentions the same advice/idea regardless of the help seeker's thoughts or previous experience\\
\hline
\multirow{4}*{Resources} & CPS & Suggests contacting CPS for help\\
 & Counseling & Suggests getting counseling\\
 & Police & Suggests contacting police and/or higher authorities\\
 & Other online services & Suggests other online services \\
\hline
\end{tabularx}
\caption{\label{tab:simplified-codes}
Counseling strategy features used to annotate conversation instances.
}
\end{table*}
}

\section{Experiment Details}
\subsection{Baseline experiments}
\label{sec:baseline-experiment-details}
All baseline models are first implemented to search the best set of parameters without incorporating any features. We have searched training batch size, learning rate, weight decay, and warm up steps for each of the BERT-family classifiers. The best working model was with DistilBERT-base-uncased sequence classifier with 16 training batch size, learning rate as $3.44\times10^{-5}$, weight decay as $3.61\times10^{-6}$, and warm up steps as 30. We also searched the optimal value of $k$ for selecting utterances in the beginning and in the end, trying various number of turns. The performance gradually improves from encoding $k=0$ turn to $k=4$ turns, and it starts decreasing from encoding $k\geq5$ turns. The number of parameters for the classifier is about 67M and training the classifier with 10 epochs takes roughly 7 minutes on NVIDIA Tesla V100 GPU with 32GB RAM. As all experiments are conducted with 10-fold cross validation, the total running time of the model with a specific input type is around 70 minutes.


\begin{table*}\scriptsize
\centering
\begin{tabularx}{0.96\textwidth}{X}
\hline
\textbf{System Message} \\
\hline \hline
You are a helpful assistant to help me understand the chat conversation between HelpSeeker and Counselor. Briefly answer questions about the conversation. + $\{$Conversation$\}$\\
\hline
\textbf{Instruction}: ``Don't answer in sentences and answer by only choosing one from the given categories''\\
\textbf{Categories}: Pre-defined feature examples described in Table \ref{tab:llm-features-design}\\
\hline
\textbf{Feature Generating Prompts}\\
\hline
$\bullet$ Help seeker's identity: ``Who is the HelpSeeker? + $\{$Instruction$\}$ + $\{$Categories$\}$''\\
$\bullet$ Perpetrator's identity: ``Who is the perpetrator? + $\{$Instruction$\}$ + $\{$Categories$\}$''\\
$\bullet$ Type of abuse: ``What is the type of the abuse or the stress? + $\{$Instruction$\}$ + $\{$Categories$\}$''\\
$\bullet$ Severity of abuse: ``What is the nature and severity of the abuse or the stress? + $\{$Instruction$\}$ + $\{$Categories$\}$''\\
$\bullet$ Help seeker's needs: ``Why does the HelpSeeker come talk to the Counselor? + $\{$Instruction$\}$ + $\{$Categories$\}$''\\
$\bullet$ Counselor's response: ``How does the Counselor help the HelpSeeker? + $\{$Instruction$\}$ + $\{$Categories$\}$''\\
$\bullet$ Counselor's strategies: ``How does the Counselor explore the issue? + $\{$Instruction$\}$ + $\{$Categories$\}$''\\
$\bullet$ What's been tried: ``What are the things that have previously done by the HelpSeeker to resolve the situation? + $\{$Instruction$\}$ + $\{$Categories$\}$''\\
$\bullet$ Counselor's advice: ``What are the things suggested by the Counselor to resolve the situation? + $\{$Instruction$\}$ + $\{$Categories$\}$''\\
$\bullet$ Help seeker's reaction: ``What is the HelpSeeker's reaction to the Counselor's suggestion? + $\{$Instruction$\}$ + $\{$Categories$\}$''\\
$\bullet$ Counselor's negative attitudes: ``Are there any indications that the Counselor hurt the HelpSeeker's feelings? + $\{$Instruction$\}$ + $\{$Categories$\}$''\\
$\bullet$ Help seeker's negative attitudes: ``Are there any indications that the HelpSeeker didn't like the chat? Consider if they are being hopeless, doubtful, denial, dissatisfied, etc. + $\{$Instruction$\}$ + $\{$Categories$\}$\\
\hline
\textbf{Prompts for Summaries}\\
\hline
$\bullet$ Plain summary: ``Summarize the conversation in 150 words.''\\
$\bullet$ Summary with stance: ``Summarize the conversation in 150 words, focusing on whether the help seeker would have felt more positive after the conversation.''\\
\hline
\textbf{Prompts for Conversation Outcome Prediction}\\
\hline
Would the help seeker have felt more positive after the conversation? Answer `0' if they would not feel more positive at all, and answer `1' otherwise.\\
\hline
\end{tabularx}
\caption{\label{tab:llm-prompt-samples}
LLM prompt design for obtaining session level features, summaries, and conversation outcome prediction.
}
\end{table*}

\subsection{Utterance-level Feature Codebook}
\label{sec:appendix-utterance-coding}
Table \ref{tab:simplified-codes} illustrates the codebook that the annotators have used for labeling utterance level features for conversation instances in $\mathcal{D}_{small}$. The column \textbf{Feature} and \textbf{Description} shows a set of fine-grained 18 classes we used for annotation and the description of each feature. In order to apply semi-supervised approach for annotating utterance level features in $\mathcal{D}_{large}$, the utterance level feature identification should be accurate, yet using a 18-class feature set does not exhibit reliable results. To this end, we categorize features into 4 groups that are described in the \textbf{Abstract Category} column. We apply this 4-class feature group to train an utterance level feature predictor model and use the model to automatically annotate $\mathcal{D}_{large}$.

\subsection{LLM prompts for session level features}
\label{sec:llm-prompts-details}

All session level features are obtained through asking one question at a time and no questions are asked as a chain. This is to minimize potential issues of ChatGPT being hallucinated by its own previous generations. Table \ref{tab:llm-prompt-samples} describes the prompts we provide to the ChatGPT model. We also illustrate prompts that are used to generate summaries about the conversation, as well as prompts that are used to evaluate the ChatGPT model's performance on conversation outcome prediction.

\subsection{Session level features to natural language explanation}
\label{sec:appendix-feat-to-explanation}

Given a set of session level features, we use a pre-defined template to convert the features into natural language explanation. We tried an alternative approach to convert features into natural language explanation by prompting ChatGPT; we prompt ChatGPT to generate explanations using a given set of features. However, the conversation outcome prediction models better fit when we use templates to convert features, thus our final method becomes using templates. Following paragraph is the template we used.

\begin{figure}[t!]
    \centering
    \includegraphics[width=1.00\columnwidth]{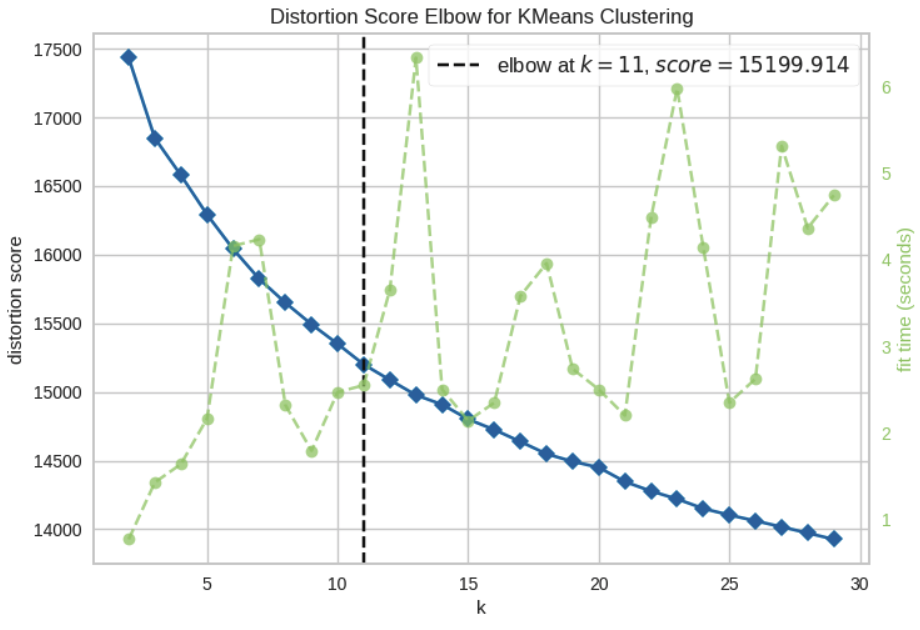}
    \caption{Distortion values of different number of clusters. Blue line indicates distortion values}
    \label{fig:inertia-values-clusters}
\end{figure}

\begin{Verbatim}[fontsize=\small,commandchars=\\\(\)]
A(n) (\color(red)[help seeker's identity]) is seeking for (\color(red)[help)
(\color(red)seeker's needs]) regarding the situation where 
there has been (\color(red)[type and severity of abuse]) by
(\color(red)[perpetrator's identity]). The counselor explores 
the issues with (\color(red)[counselor's strategies]) and 
focuses on (\color(red)[counselor's response]). The help 
seeker tried (\color(red)[what's been tried]) to resolve the 
situation and the counselor suggests (\color(red)[counselor's) 
(\color(red)advice]). About the suggestion, the help seeker is
(\color(red)[help seeker's reaction]). In the chat, the help 
seeker shows (\color(red)[help seeker's negative attitudes]). 
The counselor's attitudes seems to be (\color(red)[counselor's) 
(\color(red)negative attitudes]) in the conversation.
\end{Verbatim}

\section{Clustering results}
\label{sec:appendix-clustering}

\begin{figure*}[t!]
    \centering
    \includegraphics[width=1.00\textwidth]{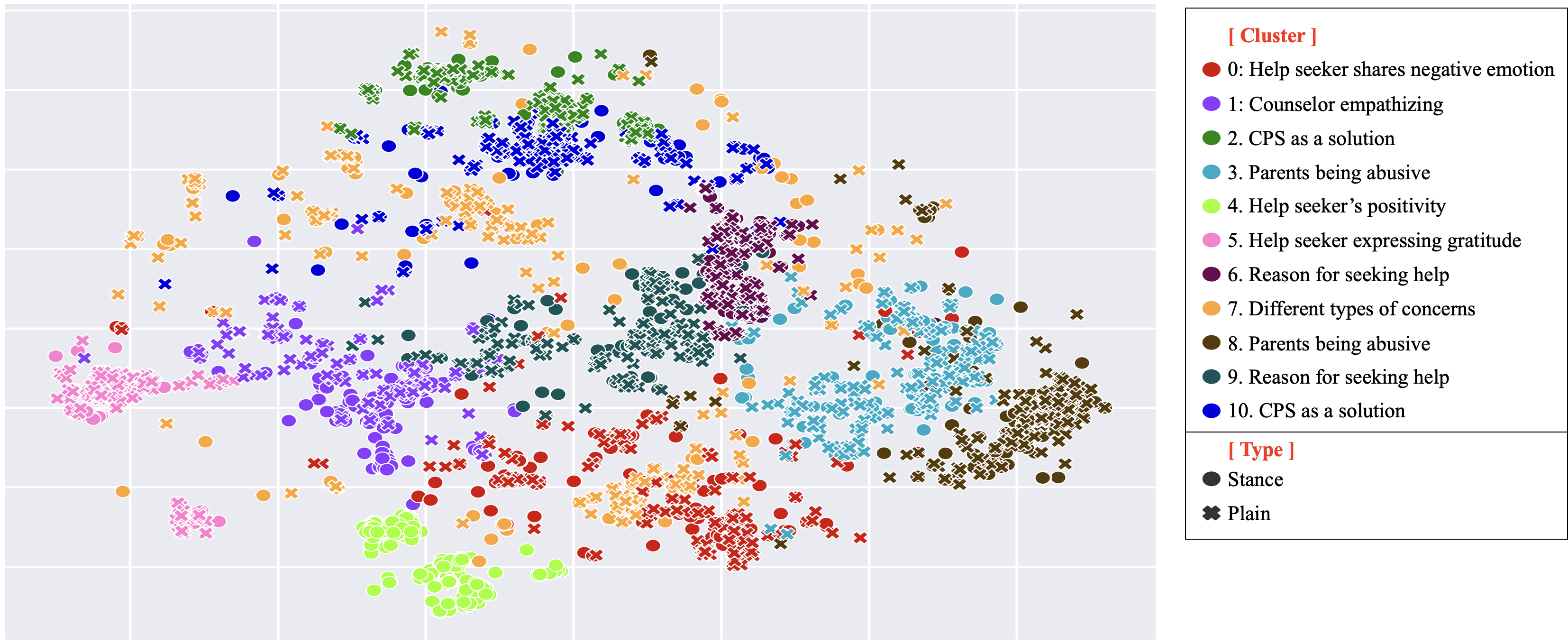}
    \caption{Clustered sentences from two types of summaries. In most case, plain summary and summary with stance produces similar aspects regarding the conversation. There are a few clusters where the portion of one summary type is meaningfully larger than the other type. Cluster 3, 5, 8 consists of around 60$\%$ of plain summary items, while cluster 1 has the opposite distribution. Cluster 4, describing the stance of the help seeker, only contains summary with stance items.}
    \label{fig:summary-clusters}
\end{figure*}

To qualitatively analyze the difference between plain summary and summary 
with stance, we perform clustering on the sentences generated by these two approaches. We first combine all summaries from the two approaches, split the sentences, encode sentences using SentenceTransformers \citep{reimers2019sentence}, and perform $k$-means clustering. The optimal $k$ is derived by comparing distortion values of different number of clusters (Figure \ref{fig:inertia-values-clusters}). 

\begin{table*}\scriptsize
\centering
\begin{tabularx}{\textwidth}{X|X}
\hline
\textbf{Cluster 0}: Help seeker shares negative emotions&\textbf{Cluster 1}: Counselor empathizing \\
\qquad\qquad\quad\verb|Summary|: $47\%$, \verb|Stance|: $53\%$&\qquad\qquad\quad\verb|Summary|: $40.67\%$, \verb|Stance|: $59.33\%$\\
\hline
$\bullet$ HelpSeeker reaches out to the Counselor, expressing their struggles with depression, anxiety, and suicidal thoughts.&$\bullet$ The Counselor provided support and empathized with the HelpSeeker's concerns.\\
$\bullet$ HelpSeeker expresses their depression and feeling of helplessness.&$\bullet$ The Counselor empathizes with the situation, reassuring HelpSeeker and offering support.\\
$\bullet$ HelpSeeker expressed feelings of sadness, wanting to end their life, and self-harm tendencies.&$\bullet$ The counselor empathizes with HelpSeeker's situation and offers support.\\
\hline
\hline
\textbf{Cluster 2}: CPS as a solution&\textbf{Cluster 3}: Parents being abusive\\
\qquad\qquad\quad\verb|Summary|: $50\%$, \verb|Stance|: $50\%$&\qquad\qquad\quad\verb|Summary|: $57.33\%$, \verb|Stance|: $42.67\%$\\
\hline
$\bullet$ The counselor provides the CPS phone number and advises HelpSeeker to explain their situation honestly.&$\bullet$ HelpSeeker explains their situation, detailing how their mother has physically abused them in the past.\\
$\bullet$ The counselor provides the CPS number and encourages HelpSeeker to contact them to document the situation.&$\bullet$ During the conversation, HelpSeeker shares concerns about their mom's physical abuse and erratic behavior.\\
$\bullet$ The counselor sympathized and encouraged HelpSeeker to contact Child Protective Services (CPS).&$\bullet$ HelpSeeker reveals that their mother is defensive about her actions, believing that she has never abused them.\\
\hline
\hline
\textbf{Cluster 4}: Help seeker's positivity&\textbf{Cluster 5}: Help seeker expressing gratitude\\
\qquad\qquad\quad\verb|Summary|: $0\%$, \verb|Stance|: $100\%$&\qquad\qquad\quad\verb|Summary|: $60\%$, \verb|Stance|: $40\%$\\
\hline
$\bullet$ It is likely that HelpSeeker felt more positive after the conversation, as they were provided with validation, guidance, and resources to seek help.&$\bullet$ The HelpSeeker expresses gratitude for the help and the conversation concludes with the Counselor offering further assistance if needed.\\
$\bullet$ Overall, it is likely that HelpSeeker would have felt more positive after the conversation due to receiving validation, resources, and a supportive response from the counselor.&$\bullet$ HelpSeeker expresses gratitude, and the conversation concludes with the Counselor encouraging HelpSeeker to reach out for further assistance if needed.\\
$\bullet$ Based on the conversation, it is likely that HelpSeeker would have felt more positive after the conversation as they received empathy, understanding, and resources for help.&$\bullet$ HelpSeeker expresses gratitude and the conversation ends on a positive note, with the counselor offering further assistance if needed.\\
\hline
\hline
\textbf{Cluster 6}: Reason for seeking help&\textbf{Cluster 7}: Different types of concerns\\
\qquad\qquad\quad\verb|Summary|: $57.33\%$, \verb|Stance|: $42.67\%$&\qquad\qquad\quad\verb|Summary|: $56.67\%$, \verb|Stance|: $43.33\%$\\
\hline
$\bullet$ HelpSeeker reached out to Counselor to discuss their concerns about being emotionally abused.&$\bullet$ HelpSeeker expresses concern and seeks advice on whether they should report the situation.\\
$\bullet$ HelpSeeker reaches out to the counselor to understand what constitutes abuse.&$\bullet$ HelpSeeker is unsure whether they should report the situation.\\
$\bullet$ HelpSeeker reached out to the Counselor seeking advice regarding their experience with child abuse.&$\bullet$ HelpSeeker asked if they could report the incident and get help.\\
\hline
\hline
\textbf{Cluster 8}: Parents being abusive&\textbf{Cluster 9}: Reason for seeking help \\
\qquad\qquad\quad\verb|Summary|: $60\%$, \verb|Stance|: $40\%$&\qquad\qquad\quad\verb|Summary|: $53\%$, \verb|Stance|: $47\%$\\
\hline
$\bullet$HelpSeeker explained that their mom constantly belittles them and their dad has physically harmed them in the past.&$\bullet$ HelpSeeker reached out to the counselor seeking advice and clarification on their parents' behavior.\\
$\bullet$ They also mentioned experiencing abuse and feeling scared of their mom.&$\bullet$ HelpSeeker reaches out to the Counselor with concerns about their mother's behavior.\\
$\bullet$ They explain that they are having issues with their family, particularly with their disrespectful mother.&$\bullet$ HelpSeeker reached out to the counselor to discuss the problems they were having with their mom.\\
\hline
\hline
\multicolumn{2}{l}{\textbf{Cluster 10}: CPS as a solution\qquad\qquad\quad\verb|Summary|: $53\%$, \verb|Stance|: $47\%$}\\
\hline
\multicolumn{2}{l}{$\bullet$ The Counselor provides guidance to HelpSeeker and suggests contacting Child Protective Services to report the situation.}\\
\multicolumn{2}{l}{$\bullet$ Counselor acknowledges HelpSeeker's concerns and suggests contacting child protective services to report the situation.}\\
\multicolumn{2}{l}{$\bullet$ Counselor advised HelpSeeker to document their observations and report the situation to Child Protective Services.}\\
\hline

\end{tabularx}
\caption{\label{tab:cluster-results}
Each cluster's topic, most representative situation examples, and the distribution of plain summary and summary with stance within the cluster.
}
\end{table*}

Figure \ref{fig:summary-clusters} illustrates clustered results after mapping sentence representations into 2d through T-distributed Stochastic Neighbor Embedding (t-SNE). The closest items to each cluster centroid and the distribution of two different summaries in each cluster are described in Table \ref{tab:cluster-results}.

\end{document}